\documentclass[letterpaper, 10 pt, journal, twoside]{ieeetran}

\IEEEoverridecommandlockouts                              %

\usepackage{graphicx} %
\usepackage{array,booktabs,arydshln}

\usepackage{cite}
\usepackage{algorithm}
\usepackage{algorithmic}
\usepackage{xcolor}

\definecolor{green}{RGB}{11,155,13}

\usepackage{amsmath, amssymb}
\usepackage{bbm}
\usepackage{diagbox}
\usepackage{makecell}
\usepackage[algo2e, ruled, linesnumbered]{algorithm2e}

\DeclareMathOperator*{\argmin}{arg\,min}
\DeclareMathOperator*{\argmax}{arg\,max}
\usepackage{multirow}

\usepackage{etoolbox}
\makeatletter
\patchcmd{\@makecaption}
  {\scshape}
  {}
  {}
  {}
\makeatother

\title{APPLE: Adaptive Planner Parameter Learning \\
from Evaluative Feedback
}

\author{Zizhao Wang$^{1}$, Xuesu Xiao$^{2}$, Garrett Warnell$^{2, 3}$, and Peter Stone$^{2, 4}$%
\thanks{Manuscript received: February, 24, 2021; Revised May, 13, 2021;
Accepted July, 12, 2021.}%
\thanks{This paper was recommended for publication by Editor Stephen J. Guy upon evaluation of the Associate Editor and Reviewers' comments. 
}%
\thanks{$^{1}$Department of Electrical and Computer Engineering,
        University of Texas at Austin, Austin, Texas 78712.
        {\tt\small zizhao.wang@utexas.edu},
        $^{2}$Department of Computer Science,
        University of Texas at Austin, Austin, Texas 78712.
        {\tt\small \{xiao, pstone\}@cs.utexas.edu},
        $^{3}$Computational and Information Sciences Directorate, Army Research Laboratory, Austin, Texas 78712. 
        {\tt\small garrett.a.warnell.civ@mail.mil},
        $^{4}$Sony AI
        }%
\thanks{This work has taken place in the Learning Agents Research Group (LARG) at the University of Texas at Austin.  LARG research is supported in part by grants from the National Science Foundation (CPS-1739964, IIS-1724157, NRI-1925082), the Office of Naval Research (N00014-18-2243), Future of Life Institute (RFP2-000), Army Research Office (W911NF-19-2-0333), DARPA, Lockheed Martin, General Motors, and Bosch. The views and conclusions contained in this document are those of the authors alone. Peter Stone serves as the Executive Director of Sony AI America and receives financial compensation for this work. The terms of this arrangement have been reviewed and approved by the University of Texas at Austin in accordance with its policy and objectivity in research.}%
\thanks{Digital Object Identifier (DOI): see top of this page.}
}

\begin{document}

\maketitle

\begin{abstract}
Classical autonomous navigation systems can control robots in a collision-free manner, oftentimes with verifiable safety and explainability. When facing new environments, however, fine-tuning of the system parameters by an expert is typically required before the system can navigate as expected. To alleviate this requirement, the recently-proposed Adaptive Planner Parameter Learning paradigm allows robots to \emph{learn} how to dynamically adjust planner parameters using a teleoperated demonstration or corrective interventions from non-expert users. However, these interaction modalities require users to take full control of the moving robot, which requires the users to be familiar with robot teleoperation. As an alternative, we introduce \textsc{apple}, Adaptive Planner Parameter Learning from \emph{Evaluative Feedback} (real-time, scalar-valued assessments of behavior), which represents a less-demanding modality of interaction. Simulated and physical experiments show \textsc{apple} can achieve better performance compared to the planner with static default parameters and even yield improvement over learned parameters from richer interaction modalities.
\end{abstract}

\section{INTRODUCTION}
\label{sec::intro}

\IEEEPARstart{M}{obile}
robot navigation is a well-studied problem in the robotics community. Many classical approaches have been developed over the last several decades and several of them have been robustly deployed on physical robot platforms moving in the real world~\cite{quinlan1993elastic, fox1997dynamic}, with verifiable guarantees of safety and explainability.

However, prior to deployment in a new environment, these approaches typically require parameter re-tuning in order to achieve robust navigation performance. For example, in cluttered environments, a low velocity and high sampling rate are necessary in order for the system to be able to generate safe and smooth motions, whereas in relatively open spaces, a large maximum velocity and relatively low sampling rate are needed in order to achieve optimal navigation performance. This parameter re-tuning requires robotics knowledge from experts who are familiar with the inner workings of the underlying navigation system, and may not be intuitive for non-expert users~\cite{zheng2021ros}. Furthermore, using a single set of parameters assumes the same set will work well on average in different regions of a complex environment, which is often not the case. 

To address these problems, Adaptive Planner Parameter Learning (\textsc{appl}) is a recently-proposed paradigm that opens up the possibility of dynamically adjusting parameters to adapt to different regions, and enables non-expert users to fine-tune navigation systems through modalities such as teleoperated demonstration~\cite{xiao2020appld} or corrective interventions~\cite{wang2021appli}.
These interaction modalities require non-expert users to take full control of the moving robot during the entire navigation task, or, at least, when the robot suffers from poor performance. However, non-expert users who are inexperienced at controlling the robot may be unwilling or unable to take such responsibility due to perceived risk of human error and causing collisions.
\begin{figure}[t]
  \centering
  \includegraphics[width=\columnwidth]{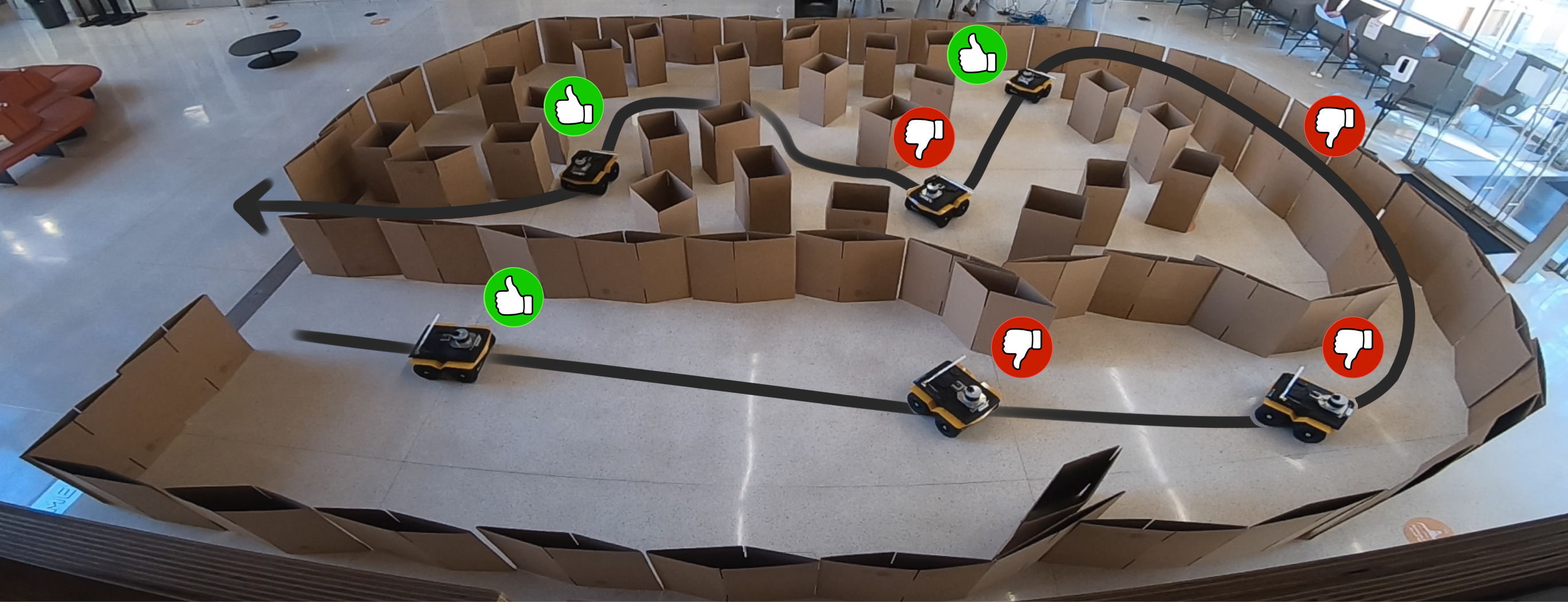}
  \vspace{-15pt}
  \caption{For non-expert users who are unable or unwilling to take control of the robot, evaluative feedback, e.g. {\em good job} (green thumbs up) or {\em bad job} (red thumbs down), is a more accessible human interaction modality, but still valuable for improving navigation systems during deployment. }
  \label{fig::apple}
  \vspace{-15pt}
\end{figure}

Fortunately, even non-expert users who are not willing to take control of the robot are typically still able to observe the robot navigating and provide real-time positive or negative assessments of the observed navigation behavior through {\em evaluative feedback}. For example, even non-expert users can know to provide negative feedback when a robot is performing poorly, e.g., getting stuck in highly constrained spaces~\cite{xiao2020toward, liu2020lifelong} or driving unnecessarily slowly in open spaces~\cite{xiao2020agile}. This more-accessible modality provides an interaction channel for a larger community of non-expert users with mobile robots (Fig. \ref{fig::apple}).

In this work, we introduce a machine learning method that can leverage evaluative feedback in the context of autonomous navigation called \emph{Adaptive Planner Parameter Learning from Evaluative Feedback} (\textsc{apple}). Based on a parameter library~\cite{xiao2020appld, wang2021appli} or a parameter policy~\cite{xu2020applr}, \textsc{apple} learns how to choose appropriate navigation planner parameters at each time step in order to adapt to different parts of the deployment environment. 
Specifically, \textsc{apple} treats the scalar human feedback as the value for the state-action pair in the Reinforcement Learning framework during training, in which the action is the parameter set to be used by the underlying navigation system. During deployment, \textsc{apple} selects the parameters to maximize the expected human feedback value. 
We implement \textsc{apple} both in the Benchmarking Autonomous Robot Navigation (\textsc{barn}) ~\cite{perille2020benchmarking} environments and also in real-world, highly constrained obstacle courses. In both training and unseen environments, \textsc{apple} is able to outperform the planner with default parameters, and even to improve over \textsc{appl} variants learned from richer interaction modalities, such as teleoperated interventions. Our experimental results indicate that evaluative feedback is a particularly valuable form of human interaction modality for improving navigation systems during deployment.

\section{RELATED WORK}
\label{sec::related}

In this section, we review existing work on machine learning for mobile robot navigation, adaptive planner parameters, and learning from human evaluative feedback. 

\subsection{Learning for Navigation}
While autonomous navigation has been studied by the robotics community for decades, machine learning approaches have recently been extensively applied to this problem as well. Xiao, et al.~\cite{xiao2020motion} presented a survey on using machine learning for motion control in mobile robot navigation: while the majority of learning approaches tackle navigation in an end-to-end manner~\cite{bojarski2016end, pfeiffer2017perception}, it was found that approaches using learning in conjunction with other classical navigation components were more likely to have achieved better navigation performance.
These methods included those that learned sub-goals~\cite{stein2018learning}, local planners~\cite{gao2017intention, chiang2019learning, xiao2020toward, xiao2020agile, liu2020lifelong}, or planner parameters~\cite{teso2019predictive, bhardwaj2020differentiable, binch2020context, xiao2020appld, wang2021appli, xu2020applr}. Learning methods have also enabled navigation capabilities that complement those provided in the classical navigation literature, including terrain-aware~\cite{wigness2018robot, siva2019robot, kahn2021badgr} and social~\cite{hart2020using, liang2020crowd, everett2018motion} navigation. 

\textsc{apple} leverages the aforementioned hybrid learning and classical architecture, where the learning component only learns to select appropriate set of planner parameters, and interacts with the underlying classical navigation system. 

\subsection{Adaptive Parameters for Classical Navigation} 
Considering classical navigation systems' verifiable safety, explainability, and stable generalization to new environments, and the difficulty in fine-tuning those systems, learning adaptive planner parameters is an emerging paradigm of combining learning and planning. Examples include finding trajectory optimization coefficients using Artificial Neural Fuzzy Inference Improvement~\cite{teso2019predictive}, optimizing two different sets of parameters for straight-line and U-turn scenarios with genetic algorithms~\cite{binch2020context}, or designing novel systems that can leverage gradient descent to match expert demonstrations~\cite{bhardwaj2020differentiable}. Recently, the \textsc{appl} paradigm~\cite{xiao2020appld, wang2021appli, xu2020applr} has been proposed, which further allows parameters to be appropriately adjusted during deployment ``on-the-fly'', in order to adapt to different regions of a complex environment. \textsc{appl} also learns from non-expert users using teleoperated demonstration~\cite{xiao2020appld}, corrective interventions~\cite{wang2021appli}, or trial-and-error in simulation~\cite{xu2020applr}. 

\textsc{apple} utilizes an accessible but sparse modality of human interaction in evaluative feedback, which is suitable for non-expert users who are not able to take control of the robot. It is also suitable for scenarios where extensive trial-and-error is not feasible and a handcrafted reward function is not available, e.g., using Reinforcement Learning (RL) \cite{xu2020applr}. Similar, or even better, navigation performance, compared to that learned from richer interaction modalities, can be achieved using \textsc{apple}.

\subsection{Learning from Human Feedback}
The method we propose in this paper uses evaluative feedback from a human to drive a machine learning process that seeks to increase the performance of an autonomous navigation system.
Because evaluative feedback is a relatively easy signal for humans to provide, several methods have been proposed to allow machines to learn from such signals over the past several decades.
Broadly speaking, most of these methods can be understood as trying to interpret the feedback signal in the context of the classical RL framework.
For example, the \textsc{coach} framework \cite{macglashan2017interactive} interprets evaluative feedback as the policy-dependent {\em advantage}, i.e., it is assumed that the feedback indicates how much better or worse the agent's current behavior is compared to what the human currently expects the agent to do.
The \textsc{tamer} framework \cite{knox2009interactively}, on the other hand, can be thought of as interpreting evaluative feedback to be the {\em value}, or expected payoff, of the current behavior if the agent were to act in the future in the way the human desires.
Yet other approaches interpret evaluative feedback directly as reward or some related statistic \cite{isbell2001social, thomaz2006reinforcement, pilarski2011online}.

\textsc{apple} adopts a similar learning from feedback paradigm, but instead of taking actions as raw motor commands, \textsc{apple}'s action space is the parameters used by the underlying navigation system. During training, \textsc{apple} learns the value of state-action pairs based on a scalar human feedback. During deployment, \textsc{apple} selects the parameters to maximize the expected human feedback.

\section{APPROACH}
\label{sec::approach}

In this section, we introduce \textsc{apple}, which has two novel features: (1) compared to learning from demonstration or interventions, which require human driving expertise in order to take full control of the moving robot, \textsc{apple} requires instead just evaluative feedback that can be provided even by non-expert users; (2) in contrast to previous work \cite{xiao2020appld, wang2021appli} that selects the planner parameter set based on how similar the deployment environment is to the demonstrated environment, \textsc{apple} is based on the expected evaluative feedback, i.e., the actual navigation performance.  
\textsc{apple}'s performance-based parameter policy has the potential to outperform previous approaches that are based on similarity.

\subsection{Problem Definition}
 
We denote a classical parameterized navigation system as $G: \mathcal{X} \times \Theta \rightarrow \mathcal{A}$, where $\mathcal{X}$ is the state space of the robot (e.g., goal, sensor observations), $\Theta$ is the parameter space for $G$ (e.g., max speed, sampling rate, inflation radius), and $\mathcal{A}$ is the action space (e.g., linear and angular velocities). During deployment, the navigation system repeatedly estimates state $x$ and takes action $a$ calculated as $a = G(x; \theta)$. Typically, a default parameter set $\bar{\theta}$ is tuned by a human designer trying to achieve good performance in most environments. However, being good at everything often means being great at nothing: $\bar{\theta}$ usually exhibits suboptimal performance in some situations and may even fail (is unable to find feasible motions, or crashes into obstacles) in particularly challenging ones~\cite{xiao2020appld}. 

To mitigate this problem, \textsc{apple} learns a parameter policy from human evaluative feedback with the goal of selecting the appropriate parameter set $\theta$ (from either a discrete parameter set library or from a continuous full parameter space) for the current deployment environment.
In detail, a human can supervise the navigation system's performance at state $x$ by observing its action $a$ and giving corresponding evaluative feedback $e$.
Here, the evaluative feedback can be either discrete (e.g., ``good/bad job") or continuous (e.g., a score ranging in $[0, 1]$).
During feedback collection, \textsc{apple} finds (1) a parameterized predictor $F_\phi: \mathcal{X} \times \Theta \rightarrow \mathcal{E}$ that predicts human evaluative feedback for each state-parameter pair $(x, \theta)$, and (2) a parameterized parameter policy $\pi_\psi: \mathcal{X} \rightarrow \Theta$ that determines the appropriate planner parameter set for the current state.

Based on whether \textsc{apple} chooses the parameter set from a library or the parameter space, we introduce the discrete and continuous parameter policies in the following two sections, respectively.

\subsection{Discrete Parameter Policy}

In some situations, the user may already have $K$ candidate parameter sets (e.g., the default set or sets tuned for special environments like narrow corridors, open spaces, etc.) which together make up a parameter library $\mathcal{L} = \{\theta^i\}_{i=1}^K$ (superscript $i$ denotes the index in the library). In this case, \textsc{apple} uses the provided evaluative feedback $e$ in order to learn a policy that selects the most appropriate of these parameters given the state observation $x$.

To do so, we parameterize the feedback predictor $F_\phi$ in a way similar to the value network in DQN \cite{mnih2015humanlevel}, where the input is the observation $x$ and the output is $K$ predicted feedback values $\{\hat{e}^i\}_{i=1}^K$, one for each parameter set $\theta^i$ in the library $\mathcal{L}$, as a prediction of the evaluative feedback a human user would give if the planner were using the respective parameter set at state $x$. We form a dataset for supervised learning, $\mathcal{D} := \{x_j, \theta_j, e_j\}_{j=1}^N$ ($\theta_j \in \mathcal{L}$, subscript $j$ denotes the time step) 
using the evaluative feedback collected so far, and $F_\phi$ is learned via supervised learning to minimize the difference between predicted feedback and the label,
\begin{equation}
    \phi^* = \argmin_\phi \mathop{\mathbb{E}}_{(x_j, \theta_j, e_j) \sim \mathcal{D}} \ell(F_\phi(x_j, \theta_j), e_j)
    \label{eq:critic_loss}
\end{equation}
where $\ell(\cdot, \cdot)$ is the categorical cross entropy loss if the feedback $e_j$ is discrete (e.g., ``good job/bad job'' or an integer score between 1 to 5),  or mean squared error given continuous feedback.

To achieve the best possible performance, the parameter policy $\pi(\cdot|x)$ chooses the parameter set that maximizes the expected human feedback (the discrete parameter policy doesn't require any additional parameters beyond $\phi$ for $F$, so the $\psi$ is omitted here for simplicity). More specifically, 
\begin{equation}
     \pi(\cdot|x) = \argmax_{\theta \in \mathcal{L}} F_{\phi^*}(x, \theta).
     \label{eqn::discrete}
\end{equation}

Compared to RL, especially DQN, discrete \textsc{apple} has a similar architecture and training objective. However, an important difference is that while RL optimizes future (discounted) cumulative reward, \textsc{apple} greedily maximizes the current feedback. The reason is that we assume, while supervising the robot's actions, the human will not only consider the current results but also future consequences and give the feedback accordingly. This assumption is consistent with past systems such as \textsc{tamer}~\cite{knox2009interactively}. Under this interpretation of feedback, \textsc{apple} can also be though of as trying to maximize some notion of future performance.

\subsection{Continuous Parameter Policy}

If a discrete parameter library is not available or desired, \textsc{apple} can also be used over continuous parameter spaces (e.g., deciding the max speed from $[0.1, 2]\ \mathrm{m/s}$). 
In this scenario, \textsc{apple} can still learn from either discrete or continuous feedback. However, learning from discrete feedback of finer resolutions or even continuous feedback should lead to better performance.

In this setting, we parameterize the parameter policy $\pi_\psi$ and the feedback predictor $F_\phi$ in the actor-critic style.
With the collected evaluative feedback $ \mathcal{D} := \{x_j, \theta_j, e_j\}_{j=1}^N$, the training objective of $F_\phi$ is still to minimize the difference between predicted and collected feedback, as specified by Eqn. (\ref{eq:critic_loss}). For the parameter policy $\pi_\psi$, beyond choosing the action that maximizes expected feedback, its training objective is augmented by maximizing the entropy of policy $\mathcal{H}(\pi_\psi(\cdot|x))$ at state $x$. Using the same entropy regularization as Soft Actor Critic (SAC)~\cite{haarnoja2018soft}, $\pi_\psi$ favors more stochastic policies, leading to better exploration during training:
\begin{equation}
     \psi^* = \argmin_\psi \mathop{\mathbb{E}}_{\substack{x_j \in \mathcal{D} \\ \tilde{\theta}_j \sim \pi_\psi(\cdot | x_j)}} \left[-F_\phi(x_j, \tilde{\theta}_j) + \alpha \log \pi_\psi(\tilde{\theta}_j | x_j)  \right],
     \label{eqn::continuous}
\end{equation}
where $\alpha$ is the temperature controlling the importance of the entropy bonus and is automatically tuned as in SAC~\cite{haarnoja2018soft}.

\subsection{Deployment}
During deployment, we measure the state $x_t$, use the parameter policy to obtain a set of parameters $\theta_t \sim \pi_{\psi^*}(\cdot | x_t)$ at each time step, and apply that parameter set to the navigation planner $G$.

\section{EXPERIMENTS}
\label{sec::experiments}
In our experiments, we aim to show that \textsc{apple} can improve navigation performance by learning from evaluative feedback, in contrast to a teleoperated demonstration or a few corrective interventions, both of which require the non-expert user to take control of the moving robot. We also show \textsc{apple}'s generalizability to unseen environments. We implement \textsc{apple} on a ClearPath Jackal ground robot in \textsc{barn}~\cite{perille2020benchmarking} with 300 navigation environments randomly generated using Cellular Automata, and in two physical obstacle courses. 

\subsection{Implementation}

The Jackal is a differential-drive robot equipped with a Velodyne LiDAR that we use to obtain a 720-dimensional planar laser scan with a 270$^\circ$ field of view, denoted as $l_t$.
The robot uses the Robot Operating System \texttt{move\textunderscore base} navigation stack with Dijkstra's global planner and the default \textsc{dwa} local planner~\cite{fox1997dynamic}.
From the global planner, we query the relative local goal direction $g_t$ (in angle) as the averaged tangential direction of the first 0.5m global path. The state space of the robot is the combination of the laser scan and local goal $x_t = (l_t, g_t)$. The parameter space consists of the 8 parameters of the \textsc{dwa} local planner as described in Tab. \ref{tab::jackal_parameters}, and the action space is the linear and angular velocity of the robot, $a_t = (v_t, \omega_t)$.

For discrete \textsc{apple}, we construct the parameter library shown in Tab. \ref{tab::jackal_parameters} with the default \textsc{dwa} parameter set $\theta_1$ and parameter sets $\theta_{2 \sim 7}$ learned in the \textsc{appli} work~\cite{wang2021appli}. Here, we use parameter sets learned in previous work, and we have found that this is important—without a reasonably good set to start with, \textsc{apple} is typically unable to learn, e.g., smaller \emph{inflation\_radius} and larger \emph{vtheta\_samples} in $\theta_2$ achieve good navigation in tight spaces, while larger \emph{max\_vel\_x} and \emph{pdist\_scale} in $\theta_4$ perform well in open ones. Without prelearned parameter sets, one can obtain a library via coarse tuning to create various driving modes (e.g. increase \emph{max\_vel\_x} to create an aggressive mode). For continuous \textsc{apple}, the parameter ranges for the parameter policy $\pi_\psi$ to select from are listed in the same table.

\begin{table}[h]
  \caption{Parameter Library and Range: \\ \emph{max\_vel\_x} \textnormal{(v)}, \emph{max\_vel\_theta} \textnormal{(w)}, \emph{vx\_samples} \textnormal{(s)}, \emph{vtheta\_samples} \textnormal{(t)}, \emph{occdist\_scale} \textnormal{(o)}, \emph{pdist\_scale} \textnormal{(p)}, \emph{gdist\_scale \textnormal{(g)}}, \emph{inflation\_radius \textnormal{(i)}}}
  \label{tab::jackal_parameters}
  \centering
  \small
  \begin{tabular}{lrrrrrrrr}
    \toprule
                & v & w & s & t & o & p & g & i \\
    \midrule
    $\theta_1$ & 0.50 & 1.57 &  6 & 20 & 0.10 & 0.75 & 1.00 & 0.30 \\
    $\theta_2$ & 0.26 & 2.00 & 13 & 44 & 0.57 & 0.76 & 0.94 & 0.02 \\
    $\theta_3$ & 0.22 & 0.87 & 13 & 31 & 0.30 & 0.36 & 0.71 & 0.30 \\
    $\theta_4$ & 1.91 & 1.70 & 10 & 47 & 0.08 & 0.71 & 0.35 & 0.23 \\
    $\theta_5$ & 0.72 & 0.73 & 19 & 59 & 0.62 & 1.00 & 0.32 & 0.24 \\
    $\theta_6$ & 0.37 & 1.33 &  9 &  6 & 0.95 & 0.83 & 0.93 & 0.01 \\
    $\theta_7$ & 0.31 & 1.05 & 17 & 20 & 0.45 & 0.61 & 0.22 & 0.23 \\
    \midrule
    $\min$ & 0.2 & 0.31 & 4  & 8  & 0.10 & 0.10 & 0.01 & 0.10 \\
    $\max$ & 2.0 & 3.14 & 20 & 40 & 1.50 & 2.00 & 1.00 & 0.60 \\
    \bottomrule
  \end{tabular}
\end{table}

Implementation-wise, for discrete \textsc{apple}, $F_\phi(x, \theta)$ is a fully-connected neural network with 2 hidden layers of 128 neurons, taking the 721 dimensional $x_t$ as input and outputting the 7 predicted feedback signals $\hat{e}_{t, 1 \sim 7}$ for $\theta_{1 \sim 7}$ respectively. Parameter policy $\pi_\psi$ uses $\epsilon$-greedy exploration with $\epsilon$ decreasing from $0.3$ to $0.02$ during the first half of the training. For continuous \textsc{apple}, $F_\phi(x, \theta)$ shares the same architecture as discrete \textsc{apple}, except for different input (concatenation of $x_t$ and $\theta_t$) and output (scalar $\hat{e}_t$). The parameter policy $\pi_\psi$ also uses the same architecture, mapping $x_t$ to $\theta_t$.

To evaluate the performance of \textsc{apple}, we use \textsc{appli} with the same parameter library in Tab. \ref{tab::jackal_parameters} upper part, \textsc{applr} with the same parameter ranges in Tab. \ref{tab::jackal_parameters} lower part, and the Default \textsc{dwa} planner as three baselines. Since \textsc{appld} does not generalize well without a confidence-based context predictor~\cite{wang2021appli} and \textsc{appli} can therefore outperform \textsc{appld}, we do not include \textsc{appld} as one of the baselines. Despite using the same library, \textsc{appli} chooses the parameter set based on the similarity between the current observation and the demonstrated environments, while discrete \textsc{apple} uses the expected feedback.

\subsection{Simulated Experiments}

\begin{figure}
  \centering
  \includegraphics[width=\columnwidth]{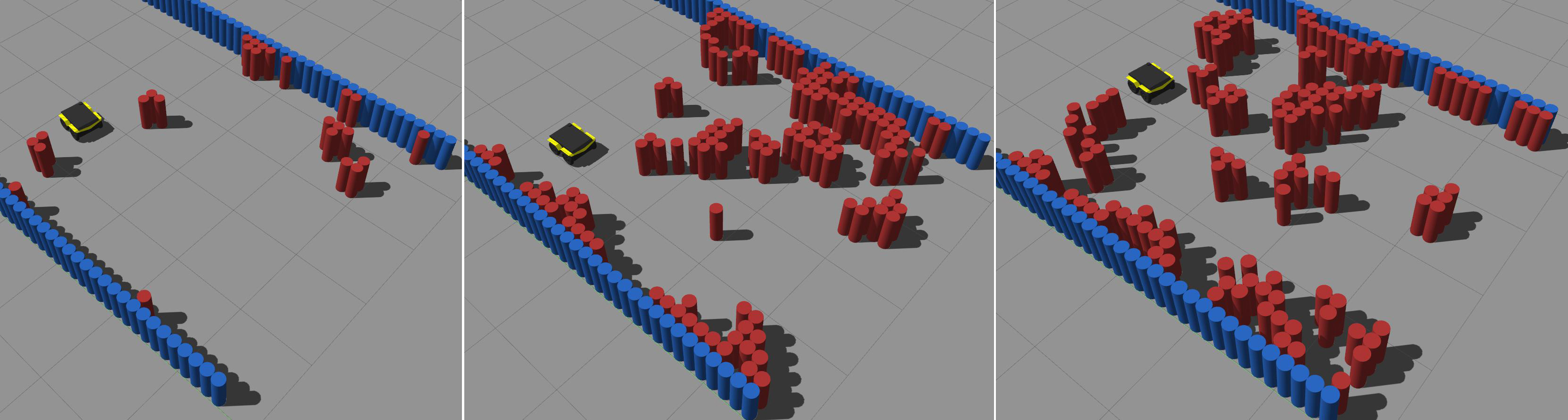}
  \vspace{-15pt}
  \caption{Simulated Environments in \textsc{barn} Dataset with Low, Medium, and High Difficulty Levels}
  \vspace{-15pt}
  \label{fig::simulated_envs}
\end{figure}

We begin by testing \textsc{apple} on the \textsc{barn} dataset, with simulated evaluative feedback generated by an oracle (proxy human). Note that although the proxy human for the simulated experiments appears to be similar to the \textsc{applr} work \cite{xu2020applr}, the simulated experiments aim to validate different \textsc{apple} setups with easily accessible feedback before physical experiments. The intended use case for \textsc{apple} is still during \emph{physical} deployments with \emph{real} humans. The benchmark dataset consists of 300 simulated navigation environments ranging from easy ones with a lot of open spaces to challenging ones where the robot needs to get through dense obstacles. Navigation trials in three example environments with low, medium, and high difficulty levels are shown in Fig. \ref{fig::simulated_envs}. We randomly select 250 environments for training \textsc{apple} and hold the remained 50 environments as the test set.

For the simulated feedback, we use the projection of the robot's linear velocity along the local goal direction, i.e., $e_t = v_t \cdot \cos(g_t)$, and it greedily encourages the robot to move along the global path as fast as possible. Then we discretize it to different number of levels ($\infty$ levels mean using continuous feedback) to study the effect of feedback resolutions. Notice, this simulated evaluative feedback is suboptimal as it doesn't consider future states, and we expect actual human evaluative feedback would be more accurate. For example, when the robot is leaving an open space and is about to enter a narrow exit, a human would expect the robot to slow down to get through the exit smoothly, but the simulated oracle still encourages the robot to drive fast. The oracle provides its evaluative feedback at 1Hz, while \textsc{apple}, \textsc{appli} and  \textsc{applr} dynamically adjust the parameter set for the \textsc{dwa} planner at the same frequency.

\begin{figure}
  \centering
  \includegraphics[width=\columnwidth]{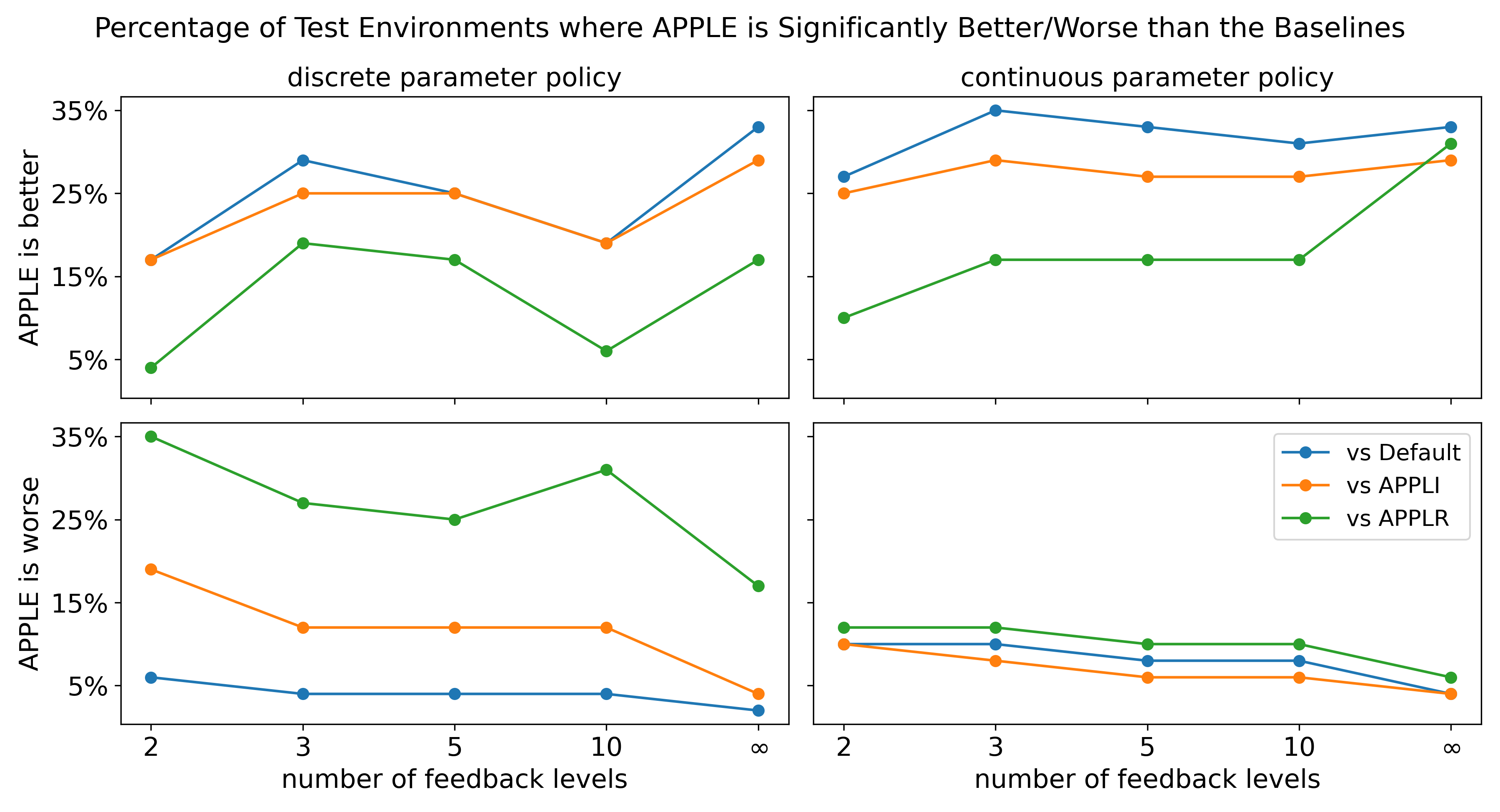}
  \vspace{-15pt}
  \caption{\textsc{apple} Learning from Different Feedback Resolutions.}
  \label{fig::apple_vs_baselines}
  \vspace{-10pt}
\end{figure}

After training in 250 environments with a total of 2.5M feedback signals collected, we evaluate \textsc{apple} with discrete and continuous parameter policies (denoted as \textsc{apple} (disc.) and \textsc{apple} (cont.)), as well as three baselines, on the 50 test environments by measuring the traversal time for 20 runs per environment. The proxy human aims at improving navigation efficiency and thus reducing traversal time.
We then conduct t-tests to compute the percentage of environments in which \textsc{apple} with different feedback resolutions is significantly better/worse ($p<0.05$) than baselines, as shown in Fig. \ref{fig::apple_vs_baselines}.

Despite learning from suboptimal evaluative feedback, \textsc{apple} (disc.) and \textsc{apple} (cont.) still outperform the Default \textsc{dwa} and \textsc{appli} at all feedback resolutions.
These results demonstrate the advantage of \textsc{apple} over \textsc{appli}, which selects the parameter set with a performance-based predictor (Eqns. \ref{eqn::discrete} and \ref{eqn::continuous}) rather than a similarity-based predictor.
Among all feedback resolutions, there is significant improvement of 3 feedback levels over 2, for both discrete and continuous policies. Further increasing resolutions doesn't improve performance much (the slight decreases are likely due to stochasticity in learning), except for using continuous feedback (i.e., $\infty$ levels) for continuous policy. These results suggest that as few as three discrete feedback levels are needed to improve navigation performance. 
Most of the cases where \textsc{apple} achieves worse performance are due to a corner case for the global planner where it keeps switching between two global paths and thus confuses the local planner. 
Surprisingly, despite that in theory \textsc{applr} is expected to perform the best, \textsc{apple} (disc.) performs only slightly worse than \textsc{applr}, and \textsc{apple} (cont.) outperforms \textsc{applr} in $29\%$ of test environments with continuous feedback. The worse performance of \textsc{applr} may come from challenging optimization of cumulative rewards or reward design. Lastly, comparing the left and right parts of Fig. \ref{fig::apple_vs_baselines}, \textsc{apple} (cont.) is comparable with \textsc{apple} (disc.) with low feedback resolutions and performs much better with continuous feedback, because of its larger model capacity in the parameter space. 

\subsection{Physical Experiments}
We also apply \textsc{apple} on a physical Jackal robot. In a highly constrained obstacle course (Fig. \ref{fig::apple}), we first apply \textsc{appli} which provides 5 sets of parameters (1 default $\theta_1$ and 4 learned $\theta_{2\sim5}$). Then to train a discrete \textsc{apple} policy which uses the same parameter library, one of the authors follows the robot autonomously navigating and uses an Xbox joystick to give binary evaluative feedback (instead of 3 feedback levels for easier collection) at 2Hz. The author aims at teaching \textsc{apple} to reduce traversal time. 
To reduce the burden of giving a large amount of feedback, the user is only requested to give negative feedback by pressing a button on the joystick when he thinks the robot's navigation performance is bad, while for other instances, positive feedback is automatically given.
In other words, we interpret the absence of human feedback to be the same as if the human had provided positive feedback.
While this interpretation is not standard in the literature (and even undesirable at times \cite{faulkner2018policy}), we found that it yielded good results for the application studied here. Because \textsc{applr} \cite{xu2020applr} requires an infeasible amount of trial and error in the real world, it is not included as a baseline in the physical experiments.

The entire \textsc{apple} training session lasts roughly 30 minutes, in which the robot navigates 10 trials in the environment shown in Fig. \ref{fig::apple}. \textsc{apple} learns in an online fashion with $30\%$ probability of random exploration. 
After the training, the learned \textsc{apple} model is deployed in the same training environment. We compare \textsc{apple} to \textsc{appli} with the same sets of parameters and the confidence measure of context prediction~\cite{wang2021appli}, and the DWA planner with static default parameters. Each experiment is repeated five times. The results are shown in Tab \ref{tab::physical}. \textsc{apple} achieves the fastest average traversal time with the smallest variance in the training environment.

\begin{figure}
  \centering
  \includegraphics[width=\columnwidth]{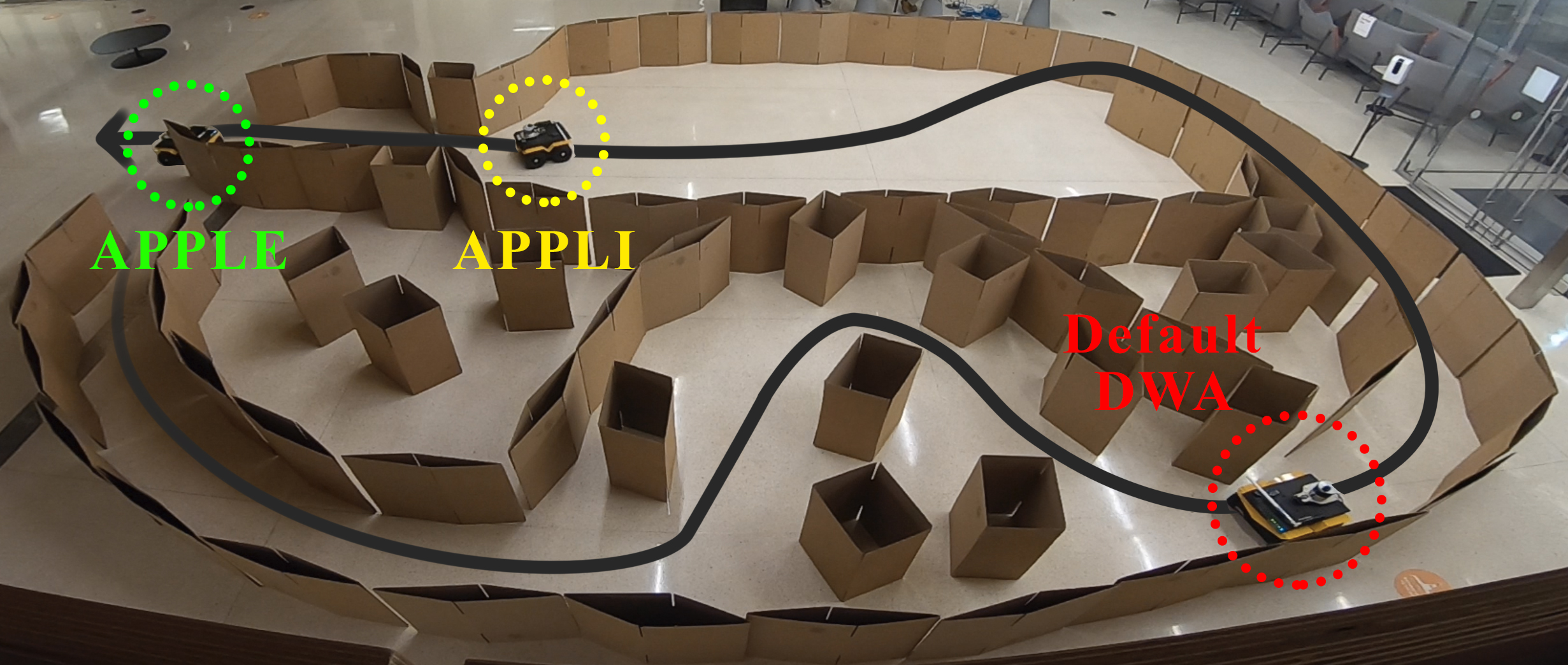}
  \vspace{-10pt}
  \caption{\textsc{apple} Running in an Unseen Physical Environment}
  \vspace{-10pt}
  \label{fig::unseen}
\end{figure}

To test \textsc{apple}'s generalizability, we also test \textsc{apple} in an unseen environment (Fig. \ref{fig::unseen}) and show the results in Tab. \ref{tab::physical}. In the unseen environment, \textsc{apple} has slightly increased variance, but still has the fastest average traversal time compared to the other two baselines.

\begin{table}
\centering
\caption{Traversal Time in Training and Unseen Environment}
\begin{tabular}{cccc}
\toprule
 & \textbf{Default} & \textbf{\textsc{appli}} & \textbf{\textsc{apple} (disc.)} \\ 
\midrule
\textbf{Training} & 143.1$\pm$20.0s & 79.8$\pm$8.1s & 75.2$\pm$4.1s \\
\textbf{Unseen} & 150.5$\pm$24.0s & 86.4$\pm$1.1s & 83.9$\pm$4.6s \\
\bottomrule
\end{tabular}
\vspace{-15pt}
\label{tab::physical}
\end{table}

\section{CONCLUSIONS}
\label{sec::conclusions}

In this work, we introduce \textsc{apple}, \emph{Adaptive Planner Parameter Learning from Evaluative Feedback}. In contrast to most existing end-to-end machine learning for navigation approaches, \textsc{apple} utilizes existing classical navigation systems and inherits all their benefits, such as safety and explainability. Furthermore, instead of requiring a full expert demonstration or a few corrective interventions that need the user to take full control of the robot, \textsc{apple} just needs evaluative feedback as simple as ``good job" or ``bad job" that can be easily collected from non-expert users. Moreover, comparing with \textsc{appli} which selects the parameter set based on the similarity with demonstrated environments, \textsc{apple} achieves better generalization by selecting the parameter set with a performance-based criterion, i.e., the expected evaluative feedback. We show \textsc{apple}'s performance improvement with simulated and real human feedback, as well as its generalizability in both 50 unseen simulated environments and an unseen physical environment. %
In this paper, we use relatively dense feedback signals from the human user in the physical experiments (and different resolutions of simulated feedback signals in the simulated experiments) to reduce the amount of time needed to train a good \textsc{apple} policy. These dense feedback signals may not always be practical, for example, the user may not always be paying attention. Therefore an important direction for future investigation is to study how little feedback is needed to yield good performance. Another important direction is to evaluate APPLE's generality with human subjects with different expertise levels and feedback criteria using an extensive user study.

\bibliographystyle{IEEEtran}
\bibliography{IEEEabrv,references}

\end{document}